# A New Network-based Algorithm for Human Activity Recognition in Videos

Weiyao Lin, Yuanzhe Chen, Jianxin Wu, Hanli Wang, Bin Sheng, and Hongxiang Li

*Abstract*—In this paper, a new network-transmission-based (NTB) algorithm is proposed for human activity recognition in videos. The proposed NTB algorithm models the entire scene as an error-free network. In this network, each node corresponds to a patch of the scene and each edge represents the activity correlation between the corresponding patches. Based on this network, we further model people in the scene as packages while human activities can be modeled as the process of package transmission in the network. By analyzing these specific "package transmission" processes, various activities can be effectively detected. The implementation of our NTB algorithm into abnormal activity detection and group activity recognition are described in detail in the paper. Experimental results demonstrate the effectiveness of our proposed algorithm.

## I. INTRODUCTION

Human activity recognition is of increasing importance in many applications including video surveillance, human-computer interaction, and video retrieval [1-14]. Automatically recognizing activities of interest plays a key part in many of the existing video systems. Therefore, it is always desirable to develop new activity recognition algorithms with higher accuracy and stronger capability for handling various scenarios.

Many algorithms have been proposed to recognize human activities [1-10]. Aggarwal and Ryoo [1] gave a comprehensive review of human activity analysis. Nascimento et al. [2] detected human actions using a bank of switch dynamical models with a *priori* knowledge of the scenario. Rao et al. [3] introduced view-invariant dynamic time warping for analyzing activities with trajectories. Zelniker et al. [4] created global trajectories by tracking people across different cameras and detected abnormal activities if the current global trajectory deviates from the normal paths. However, these algorithms only focus on recognizing the "scene-related" activities (i.e., activities only considering the relationship between the person and his surrounding scene, such as a person following a regular path or a person entering unusual regions). Thus, it is very difficult to extend these algorithms into the recognition of group activities (i.e., activities including the interaction among people such as "approach" or "people being followed" [12]). Furthermore, Kim and Grauman [5] proposed to use a Mixture of Probabilistic Principal Component Analyzers (MPPCA) to learn normal patterns of activities and infer a space-time Markov Random Field (MRF) to detect abnormal activities. This method can effectively detect and locate abnormal activities which deviate from the learned normal motion patterns and has the potential to be extended to detect group activities when the group motion pattern is suitably learned. However, since this method is constructed based on the local-region motion information, it cannot explicitly differentiate activities with motion patterns in common (e.g., differentiating moving-back-and-forth from moving-forward and moving-backward). Also, the step of inferring the MRF during the detection process is also time-consuming.

There are also a variety of researches on group activity recognition. Zhou et al. [13] detected pair-activities by extracting the causality features from bi-trajectories. Ni et al. [14] further extended the causality features into three types including individuals, pairs and groups. Cheng et al. [11] used the Group Activity Pattern for representing and differentiating group activities where Gaussian parameters from trajectories were calculated from multiple people. Lin et al. [12] used group representative to represent each group of people for detecting the interaction of people groups such that the number of people can vary in the group activity. However, while these methods suitably handle the interaction among people, many of them neglect the relationship between people and their surrounding scene. Thus, they may have limitations when detecting the scene-related activities. Furthermore, their abilities for detecting complex activities (such as people first approach and then split) are also limited.

Although some methods [7, 10, 17, 27] can recognize the group activity as well as the scene-related activity by using some pre-designed graphical models such as the layered Hidden Markov Model (HMM) [10], they often require large amount of training data in order to work well. Besides, the restricted graphical structure used in these methods may also limit their ability to handle various unexpected cases.

In this paper, we propose a new network-transmission-based (NTB) algorithm for human activity recognition. The proposed framework first models the entire scene as an error-free network. In this network, each node corresponds to a patch of the scene and each edge represents the activity correlation between the corresponding patches. Based on this network, we further model people in the scene as packages and human activities can be viewed as the process of package transmission in the network. By suitably analyzing these specific package transmission processes, human activities can be efficiently recognized. Our NTB algorithm is flexible and capable of handling both the

W. Lin and Y. Chen are with the Department of Electronic Engineering, Shanghai Jiao Tong University, Shanghai 200240, China (e-mail: {wylin, yzchen0415}@sjtu.edu.cn).

J. Wu is with the National Key Laboratory for Novel Software Technology, Nanjing University, Nanjing 210023, China (email: wujx2001@nju.edu.cn).

H. Wang is with the Department of Computer Science and Technology, Tongji University, Shanghai 201804, China (email: hanliwang@tongji.edu.cn).

B. Sheng is with the Department of Computer Science, Shanghai Jiao Tong University, Shanghai 200240, China (e-mail: shengbin@cs.sjtu.edu.cn).

H. Li is with the Department of Electrical and Computer Engineering, University of Louisville, Louisville 40292, USA (email: hongxiangli@gmail.com).

interactions among people (group activities) and the interaction between people and the scene (scene-related activities). Experimental results demonstrate the effectiveness of our proposed algorithm.

The rest of the paper is organized as follows: Section II describes the framework of our proposed NTB algorithm. Section III and Section IV describe the implementations of our NTB algorithm in abnormal event detection and group activity recognition in detail, respectively. The experimental results are shown in Section V and Section VI concludes the paper.

## II. FRAMEWORK OF THE NTB ALGORITHM

### A. Basic idea of the algorithm

The basic idea of our network-transmission-based (NTB) algorithm can be described by Fig. 1 and Fig. 2. Our NTB algorithm first divides the entire scene into patches where each patch is modeled as a "node" in the error-free network (as in Fig. 1). Based on this network, the process of people moving in the scene can be modeled as the package transmission process in the network (i.e., a person moving from one patch to another can be modeled as a 'package' transmitted from one node to another). In this way, various human activity recognition problems can be transferred into the package transmission analysis problem in the network.

With this network-based model, one key problem is how to use this model for recognizing activities. We further observe that if we model the process of person moving among patches as the 'energy' consumed to transmit a package, the activities can then be detected with these 'transmission energy' features. For example, abnormal activities can be detected if its energies deviate from the "normal activity transmission energy" by larger than a pre-trained threshold (i.e., for example, if a person moves to an unusual patch, the 'energy' used will be increased and this will be detected as an abnormal activity). In this way, the abnormal detection problem can be modeled as the energy efficient transmission problem in a network [19].

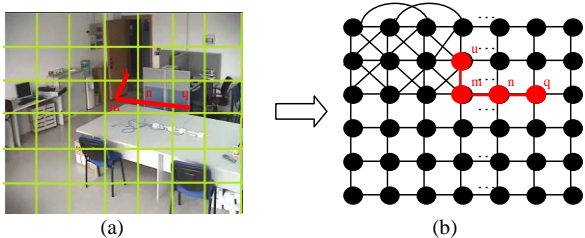

Fig. 1 (a) Divide the scene into patches. (b) Model each patch in (a) as a node in the network and the edges between nodes are modeled as the activity correlation between the corresponding patches. The red trajectory $R(u, q)$ in (a) is modeled by the red package transmission route in (b). (Note that (b) can be a fully connected network (i.e., each node has edges with all the other nodes in the network). In order to ease the description, we only draw the four neighboring edges for each node in the rest of the paper) (best viewed in color).

Furthermore, besides modeling the correlation between the person and the scene, our network-based model can also be easily extended for handling the interaction among people. For example, as in Fig. 2, we can construct a "relative" network where one person is always located in the center of the network and the movement of another person can be modeled as the package transmission process in this "relative" network based on his relative movement to the network-center person. In this way, the interaction among people can also be effectively recognized by evaluating different transmission energies in our network-based model.

Based on the above discussions, we can propose our NTB algorithm. The framework of our algorithm is described in the following section.

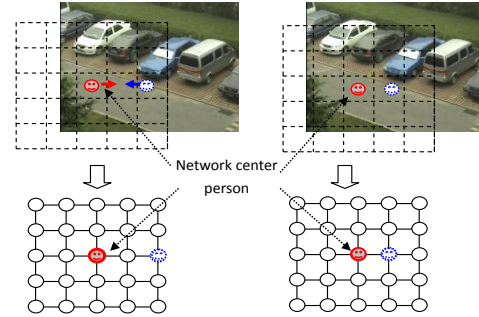

Fig. 2 Constructing relative networks for modeling people interactions. Upper: the locations of the two approaching people in two different frames (the dashed patches are divided by making the red-circled grey person at the network center). Down: the transferred networks of the upper frames (the red-circled grey node and the blue-circled dotted node are the locations of the two people in the network). The location of the red-circled grey person is fixed in the bottom network while the location of blue-circled dotted person in the bottom network is decided by his relative location to the red-circled grey person.

### B. The framework

The framework of our proposed NTB algorithm is shown in Fig. 3. In Fig. 3, the part in the dashed rectangle is the training module while the three blocks on the top are the testing process.

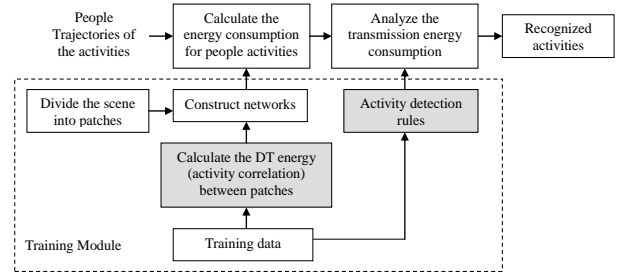

Fig. 3 The framework of the NTB algorithm.

In the training module, the scene is first divided into patches where each patch is modeled as a node in the network. Then the activity correlations between patches are estimated based on the training data and these activity correlations will be used as the edge values in the network. With these nodes and edges, the transmission networks can be constructed. At the same time, the activity detection rules are also derived from the training data for detecting activities of interest during the testing process.

In the testing process, after obtaining trajectories of people (which represent activities), their corresponding transmission energies are first calculated based on the constructed network. Then, these transmission energies are analyzed and the activity detection rules will be applied for detecting the activities.

Furthermore, several things need to be mentioned about our NTB algorithm. They are described in the following:
(1) We assume the networks used in our algorithm are error-free (i.e., there are no interferences such as noises or package losses

during transmission).

(2) Although there are other works [7, 15, 17] trying to segment the scene into parts for activity recognition, our NTB algorithm is different from them in: (a) our NTB algorithm construct a package transmission network over the patches while other works [7, 17] use graphical models for recognition. While the fixed structures of the graphical models [7, 17] may limit their ability to handle various unexpected cases, our fully-connected transmission network is more general and flexible for handling various scenarios; (b) With the transmission network model, our NTB algorithm is robust to the patch segmentation styles (e.g., in this paper, we just simply segment the scene into identical rectangular blocks as shown in Fig. 1). Comparatively, the graphical model-based methods normally require careful segmentation of the scene [15, 17].

(3) From Fig. 3, it is clear that the steps of "calculate the energy between patches" and "activity detection rules" are the key parts of our NTB algorithm. The implementation of these steps can be different for different activity recognition scenarios. Therefore, in the next two sections, we will describe the implementations of our algorithm in two scenarios (abnormal event or scene-related activity detection, and group activity recognition), respectively.

### III THE IMPLEMENTATION OF NTB ALGORITHM IN ABNORMAL EVENT DETECTION

In this scenario, we try to detect abnormal activities such as people following irregular paths and people that move back and forth. In the following, we will describe the implementation of each step in Fig. 3 in detail. Again, note that the two grey blocks in Fig. 3 are the key parts of our algorithm.

*A. Divide the scene into patches*

For the ease of implementation, we simply divide the scene into identical non-overlapping rectangular patches in this paper, as in Fig. 1. Note that other semantic-based segmentation methods [15] can also be easily used in our algorithm.

*B. Calculate the energy consumption for people activities*

Let $R(u, q)$ be the person trajectory of the current activity with $u$ being the starting patch and $q$ being the person's current patch. Also define the Direct Transmission (DT) energy for the edge between patches $i$ and $j$ as $e(i, j)$ (i.e., the energy used by directly transmitting a package from patch $i$ to $j$ without passing through other patches, as will be described in detail in the next sub-section). The total transmission energy for the trajectory $R$ can be calculated by accumulating the DT energies of all patch pairs in the trajectory, as in Eqn. (1).

$$E(u,q) = \sum_{(i,j) \in R(u,q)} e(i,j) \quad (1)$$

For example, in Fig. 1 (a), the total transmission energy for the red trajectory $R(u, q)$ equals to $e(u, m)+e(m, n)+e(n, q)$.

*C. Calculate the DT energy (activity correlation) between patches*

The edges between nodes in the network are modeled by the Direct Transmission (DT) energy between patches. In order to calculate the DT energy, we first introduce the activity correlation $AC$ between patches. That is, when there are high chances for people to perform activities between two patches $i$ and $j$ (e.g., move across $i$ and $j$), a high correlation $AC(i, j)$ will appear between these patches. Otherwise, a low correlation will be set. Thus, the activity correlation can be calculated by:

$$AC(i,j) = \sum_k tw_k(i,j) \quad (2)$$

where $AC(i, j)$ is the activity correlation between patches $i$ and $j$. $tw_k(i, j)$ is the correlation impact weight between $i$ and $j$ from the $k$-th trajectory in the training data. From Eqn. (2), we can see that the activity correlation $AC(i, j)$ is the summation of correlation impact weights $tw_k(i, j)$ from the training trajectories. If more training trajectories indicate a high correlation between patches $i$ and $j$, a large activity correlation $AC(i, j)$ will be calculated. With the definition of $AC(i, j)$, the DT energy $e(i, j)$ between patches can be calculated by:

$$e(i,j) = 1/AC(i,j) \quad (3)$$

From Eqn. (3), we can see that the DT energy is inversely proportional to the activity correlation. That is, when the activity correlation value $AC(i, j)$ are larger between patches $i$ and $j$, it implies that a "higher" activity correlation will appear between the patches, resulting in a "lower" DT energy. In this way, we can guarantee that normal activities (normally go across high-correlation patches) can result in smaller total energies.

From Eqns (2)-(3), we can see that the correlation impact weights $tw_k(i, j)$ are the key parts for calculating the DT energies. In this paper, an iterative method is proposed to calculate $tw_k(i, j)$, $AC(i, j)$, and $e(i, j)$ and the flowchart of this iterative method is shown in Fig. 4.

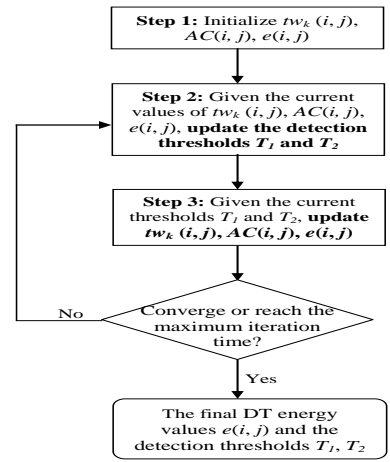

Fig. 4 The flowchart of the iterative method.

From Fig. 4, we can see that the proposed iterative method mainly includes three steps: In the first step, $tw_k(i, j)$, $AC(i, j)$, and $e(i, j)$ are initialized. In the second step, given the current values of $tw_k(i, j)$, $AC(i, j)$, and $e(i, j)$, the thresholds for activity detection ($T_1$ and $T_2$) are updated such that they can achieve good detection results with the current DT energy values $e(i, j)$. In the third step, the values of $tw_k(i, j)$, $AC(i, j)$, $e(i, j)$ are further updated with the newly updated detection thresholds ($T_1$ and $T_2$).

And step 2 and step 3 will be performed iteratively until the parameter values are converged or the maximum iteration time is reached. From Fig. 4, we can see that the key parts of the iterative method are the three steps. Therefore, in the following, we will describe the detailed process of the three steps in Fig. 4.

**Step 1: Initialize $tw_k(i, j)$, $AC(i, j)$, and $e(i, j)$**. The values of $tw_k(i, j)$, $AC(i, j)$, and $e(i, j)$ are initialized by:

$$tw_k^0(i,j) = \begin{cases} 1 & \text{if training trajectory } k \text{ moves across patch } i \text{ and } j \\ 0 & \text{otherwise} \end{cases} \quad (4)$$

$$AC^0(i,j) = \sum_k tw_k^0(i,j) \text{ and } e^0(i,j) = 1/AC^0(i,j)$$

where $tw_k^0(i, j)$, $AC^0(i, j)$, and $e^0(i, j)$ are the initialized values and the superscript 0 stands for the iteration number. From Eqn. (4), we can see that $tw_k(i, j)$ is initialized to be 1 if the $k$-th training trajectory moves across node $i$ and node $j$, or initialized to be 0 otherwise. This means that the initial DT energy values $e^0(i, j)$ are set to be the inverse of the total number of training trajectories crossing the patches such that a large number of crossing trajectories implies a high correlation between the patches and thus a low DT energy value will be initialized. This process can reasonably initialize the DT energy values close to their optimal ones. Furthermore, note that a large value will be set for initializing $e^0(i,j)$ if the trajectory-crossing time between $i$ and $j$ is zero in order to avoid dividing by 0.

**Step 2: Update the detection thresholds $T_1$ and $T_2$.** In this step, the detection thresholds $T_1$ and $T_2$ are updated such that the updated thresholds can achieve good detection results with the current DT energy values $e(i, j)$. The thresholds are updated by Eqn. (9) and this step will be described in detail in the next sub-section (Section III-D).

**Step 3: Update $tw_k(i, j)$, $AC(i, j)$, and $e(i, j)$.** Given the newly updated detection thresholds $T_1$ and $T_2$, the values of $tw_k(i, j)$, $AC(i, j)$, and $e(i, j)$ can be updated by:

$$tw_k^{n+1}(i,j) = \begin{cases} tw_k^n(i,j) & \text{if k-th trajectory is correctly recognized} \\ tw_k^n(i,j) \cdot \left(1 + \frac{E_k^n - T_l^n}{E_k^n}\right) & \text{if k is a false alarm} \\ tw_k^n(i,j) \cdot \left(1 - \frac{T_l^n - E_k^n}{2 \cdot T_l^n}\right) & \text{if k is a miss detection} \end{cases} \quad (5)$$

$$AC^{n+1}(i,j) = \sum_k tw_k^{n+1}(i,j) \text{ and } e^{n+1}(i,j) = 1/AC^{n+1}(i,j)$$

where the superscript $n$ and $n+1$ are the iteration numbers and the subscript $k$ is the trajectory number. $tw_k^{n+1}(i, j)$, $AC^{n+1}(i, j)$, and $e^{n+1}(i, j)$ are the updated values in the $n+1$ iteration. $E_k^n$ is the total transmission energy for trajectory $k$ calculated by Eqn. (1) based on the DT energy values in the $n$-th iteration $e^n(i, j)$. $T_l^n$ ($l$=1 or 2) are the detection thresholds in the $n$-th iteration where the selection of $l$ depends on which threshold is used for detection. The calculation of $T_l^n$ ($n$=0, 1, 2 ...) will be described in detail in the next subsection (Section III-D).

From Eqn. (5), we can see that during each iteration, the activity detection results on the training data are used as the feedback for updating the DT energies. In this way, a false alarm will increase the activity correlation weight $tw_k(i, j)$ and a miss detection will decrease $tw_k(i, j)$. More specifically, if trajectory $k$ is a false alarm (i.e., a normal activity wrongly detected as abnormal) and it passes through patches $i$ and $j$, we will increase $tw_k(i, j)$ according to Eqn. (5) (i.e., increase $tw_k(i, j)$ by multiplying a factor of $1+(E_k^n-T_l^n)/E_k^n$). In this way, the DT energy $e(i, j)$ will be decreased such that the total transmission energy for $k$ will become smaller in the next iteration (note that $e(i, j)$ is inversely proportional to $tw_k(i, j)$), making $k$ more likely to be detected as a normal activity. On the contrary, if the abnormal activity $k$ is detected as a normal activity (i.e., a miss), the correlation impact weight of $k$ will be decreased (or the DT energy $e(i, j)$ is increased) for increasing the ability to detect abnormal activities.

Furthermore, note that since our network model allows the nodes to be fully connected to each other (i.e., each node can also have edge with non-adjacent patches), our NTB algorithm is more general and flexible of handling unexpected cases such as a person unexpectedly 'jumping' to a non-adjacent patch due to the occlusion in the adjacent patches.

*D. Activity detection rules*

As mentioned, the basic idea of using our NTB algorithm for abnormal activity detection is to evaluate whether the 'transmission energy' of the activity deviates from the normal case by larger than a pre-trained threshold. Therefore, one of the key issues of our algorithm is to estimate the energy consumption for normal activities such that it can be used as the reference for abnormality detection. In this paper, we propose to create the minimum-energy route map for estimating the 'total transmission energies' needed for normal activities. Based on this minimum-energy route map, criteria can be developed to detect abnormal activities.

The minimum-energy route map can be described by:

$$map_{normal} = \{E_{min}(m,n), R_{min}(m,n) / m, n \in S\} \quad (6)$$

where $S$ is the entire set of all patches, $E_{min}(m, n)$ and $R_{min}(m, n)$ are the smallest possible transmission energy and its corresponding minimum energy consumption route when we want to transmit packages from patch $m$ to $n$, respectively.

Since in practice, the number of entrance (or exit) patches in the scene is limited, we do not need to calculate $E_{min}$ and $R_{min}$ for all $(m, n)$ pairs. Instead, $map_{normal}$ only need to include $E_{min}(u, n)$ and $R_{min}(u, n)$ where $u$ are the entrance (or exit) patches and $n$ is any patch in the scene (i.e., $n \in S$). In this way, given any patch in the scene, we can know the best route and its corresponding minimum transmission energy from the entrance $u$ to this patch.

Since the calculation of $E_{min}(u, n)$ and $R_{min}(u, n)$ are similar to the energy routing problem in network broadcasting [19], they can be calculated by the energy-efficient-routing algorithms [19] used in wireless sensor broadcasting. However, since we only need to calculate the minimum energy and route to the entrance patch $u$ instead of between any patches (i.e., $u$ is fixed in $E_{min}(u, n)$ and $R_{min}(u, n)$), the routing algorithm can be simplified. Therefore, in this paper, we use a Simplified Broadcast Incremental Power (SBIP) algorithm for creating the normal transmission route map. It is described as in Algorithm 1. Fig. 5 shows the process of the SBIP algorithm in an example network.

**Algorithm 1** The SBIP Method
**Input:** an undirected weighted complete graph $G(N, V)$, where $N$ is the set of its nodes and $V=\{e(i, j)\}$ is the set of its edges. The DT energy of edge from $N(i)$ to $N(j)$ is stored in $e(i, j)$. $NE$ is the set of nodes which have already been added to the routing tree. And $u$ is the start node of the route tree.
**Output:** minimum-energy route tree node set $R_{min}$ and accumulative energy vector $E_{min}$.
**Initialization:** Set all elements of $E_{min}(i, j)$ $(i \neq j)$ to a large number $L$. Also set $R_{min}(u, u)=\{u\}$; $E_{min}(u, u)=0$; $NE=\{u\}$ where $u$ is the start node
**while** $|NE| \neq |N|$ **do**
    $Ec=L$
    **for** $i=1: |N|$ and $i \in NE$
        **for** $j=1:|N|$ and $j \notin NE$
        $E\_tmp=E_{min}(u, i)+ e(i, j)$
        **if** $E\_tmp < Ec$
            $Ec=E\_tmp$; $ic= i$; $jc =j$;
        **end**
        **end**
    **end**
    **set** $E_{min}(u, jc)=Ec$; $R_{min}(u, jc)=\{R_{min}(u, ic), jc\}$;
    **add** node $jc$ to $NE$;
**end**

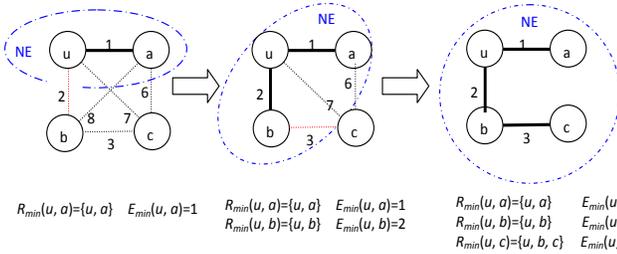

Fig. 5 The process of the SBIP algorithm in an example network. (The nodes inside the blue dash-dot circle are the set of nodes which have already been added to the routing tree (i.e., NE in Algorithm 1); The dashed lines are the DT energy values; The bold solid lines are the minimum energy routes in the tree; And the lists at the bottom are the decided $R_{min}$ and $E_{min}$ in each step)

Based on the minimum-energy route map, detection rules can then be developed to decide whether the input activity trajectory is abnormal. The proposed abnormal detection criteria are:

*The current activity $R(u, q)$ is abnormal if:*
$E(u, q) > T_1$ or $E(u, q) > T_2(u, q)$     (7)

where $R(u, q)$ is the trajectory of the current activity with $u$ being the entrance patch and $q$ being the current patch. $E(u, q)$ is the total transmission energy for the current activity and $E_{min}(u, q)$ is the minimum possible energy between $u$ and $q$ and it is calculated by Algorithm 1 and Eqn. (6). $T_l$ ($l=1$ or $2$) are the thresholds for detecting abnormal activities. Note that $T_1$ is a constant value for all trajectories while $T_2(u, q)$ is adaptive with the trajectories and controlled by the parameter $\alpha$ by:

$$T_2(u,q) = \alpha \cdot (E_{min}(u,q)) \quad (8)$$

Note that $T_1$ and $T_2$ can be automatically determined by the training data during the same recursive training process as in Fig. 4 where in each iteration, $T_1$ and $T_2$ are updated by finding a suitable set $T_1^n$ and $T_2^n$ that minimize the squared summation of two error rates $err_{FA}^2 + err_{miss}^2$:

$$\{T_1^n, T_2^n\} = \arg\min_{\{t_1, t_2\}} \left( err_{FA}\left(t_1, t_2, \{e^n(i,j)\}\right)^2 + err_{miss}\left(t_1, t_2, \{e^n(i,j)\}\right)^2 \right) \quad (9)$$

where $\{e^n(i, j)\}$ is the DT energy set for all edges in the transmission network during the $n$-th iteration (as calculated by Eqn. (5)). $err_{FA}(t_1, t_2, \{e^n(i, j)\})$ and $err_{miss}(t_1, t_2, \{e^n(i, j)\})$ are the false alarm and miss detection rates [12] for detecting abnormal activities in the training set when the thresholds $t_1$, $t_2$ as well as $\{e^n(i,j)\}$ are used for detection. It should be noted that the rules in Eqn. (7) are only one way to detect abnormal activities. In practice, other general classifiers (such as the Support Vector Machine (SVM) [8]) can also be used to take the place of Eqn. (7) and to perform abnormal activity recognition based on our energy features. And this will be further discussed in Section V.

Furthermore, Fig. 6 shows the detailed detection processes for 4 example activity trajectories. From Eqn. (7) and Fig. 6, we can see that with our detection rules, an activity will be detected as abnormal if its total energy deviates from the "normal activity transmission energy" by larger than a pre-trained threshold $T_1$ (such as Fig. 6 (b) and (d)), or it is larger than another pre-trained threshold $T_2$ (such as Fig. 6 (c)). Note that the second detection criterion is included such that: (a) The on-the-fly (or online) abnormal detection is enabled such that we can detect normal/abnormal activities in the current patch rather than waiting until the end of the trajectory for detection; (b) Some abnormal activities with small absolute energy values but large ratios over $E_{min}(u, q)$ can be effectively detected (as Fig. 6 (c)).

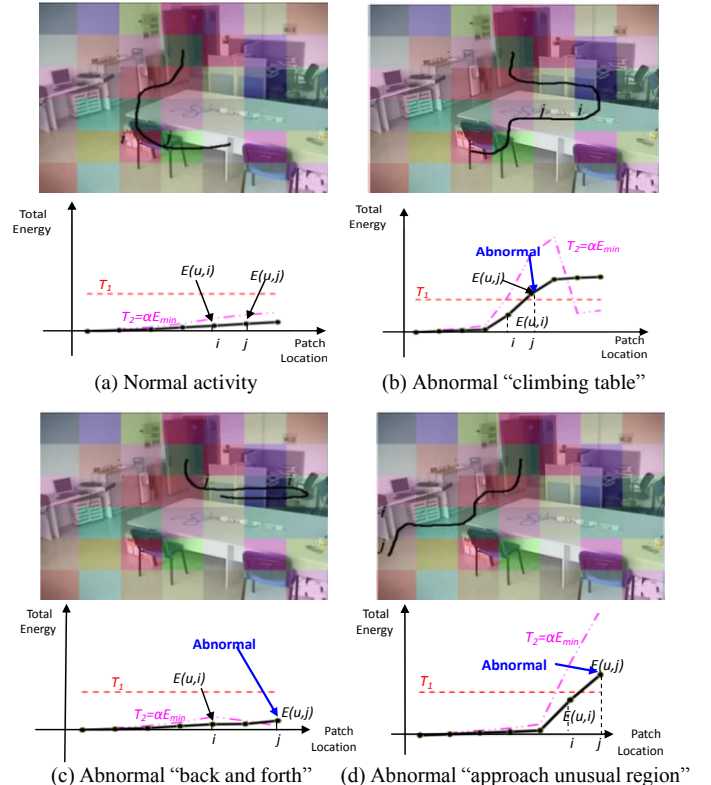

(a) Normal activity      (b) Abnormal "climbing table"
(c) Abnormal "back and forth"      (d) Abnormal "approach unusual region"
Fig. 6 Examples of the activity trajectories (upper) and their corresponding detection processes (down). (The red dashed line and the pink dash-dot lines are thresholds $T_1$ and $T_2$. The black circle-marker line is the total transmission energy. And the blue arrows are the patches where abnormalities are detected)

## IV THE IMPLEMENTATION OF NTB ALGORITHM IN GROUP ACTIVITY RECOGNITION

In the group activity recognition scenarios, we want to recognize various group activities such as people approach each other, one person leaves another, and people walk together. As mentioned, when recognizing the interaction among people, the relative networks can be constructed as in Fig. 2. At the same time, since some group activities also include the relationship between people and their surrounding scene (e.g., we need to recognize whether a person is moving or standing still in the scene in order to differentiate activities such as both people walk to "meet" or one person stand still and another one "approaches" him), a "scene-related" transmission network similar to abnormal event detection is also required. Therefore, in this section, we propose to use two types of networks for representing group activities. The detailed implementation of the key parts in Fig. 3 for group activity detection is described in the following.

*A. Construct networks*

In this paper, we construct three networks for recognizing group activities: the scene-related network, the normal relative network, and the weighted relative network. The scene-related network is used to model the correlation between people and the scene and it can be constructed as in Fig. 1. The normal relative network and the weighted relative network are used for modeling the interaction among people and they can be constructed by fixing the location of one person in the network and derive the locations of other people based on their relative movements to the location-fixed person, as in Fig. 2. Besides, the following points need to be mentioned about the networks.

(1) The structures of the normal relative and the weighted relative networks are the same. They only differ in edge values.
(2) Note that the scene-related network is an undirected network (i.e., the DT energy consumption when moving from patch $i$ to $j$ is the same as moving from $j$ to $i$). However, the normal relative network and the weighted relative network are directed networks (i.e., the DT energy from $i$ to $j$ is different from $j$ to $i$). This point will be further described in detail in the following sub-sections.
(3) Since the relative networks only focus on the relativity between people, when constructing relative networks, we randomly select one person to be the reference person and put him at the center of the relative network.
(4) Besides the three networks used in the section, our algorithm can also be extended to include other networks with other motion features. And this point will be further discussed in detail in the experimental results (i.e., Section V-C and Section V-D).

*B. Calculate the energy consumption for people activities*

In this paper, we propose to calculate a set of transmission energies from the three networks for describing group activities. For the ease of description, we use two-people group activity as the example to describe our algorithm. Multiple people scenarios can be easily extended from our description. The total transmission energy set for two-people group activity can be calculated by:

$$[E_1(u_1,q_1), E_2(u_2,q_2), ENR(u_2-u_1,q_2-q_1), EWR(u_2-u_1,q_2-q_1)] \quad (10)$$

where $E_1(u_1,q_1)$ and $E_2(u_2,q_2)$ are the total transmission consumption for person 1 and person 2 in the "scene-related" network, respectively. And they can be calculated by Eqn. (1). $ENR(u_2-u_1,q_2-q_1)$ is the total transmission consumption in the "normal relative" network where $R(u_2-u_1,q_2-q_1)$ is the relative trajectory of person 2 with respect to person 1. And $EWR(u_2-u_1,q_2-q_1)$ is the total transmission consumption in the "weighted relative" network. $ENR(u_2-u_1,q_2-q_1)$ and $EWR(u_2-u_1,q_2-q_1)$ can be calculated by:

$$\begin{cases} ENR(u_2-u_1,q_2-q_1) = \sum_{(i,j)\in R(u_2-u_1,q_2-q_1)} enr(i,j) \\ EWR(u_2-u_1,q_2-q_1) = \sum_{(i,j)\in R(u_2-u_1,q_2-q_1)} ewr(i,j) \end{cases} \quad (11)$$

where $enr(i,j)$ and $ewr(i,j)$ are the Direct Transmission (DT) energy from patch $i$ to $j$ in the normal relative network and weighted relative network, respectively. The calculation of $enr(i,j)$ and $ewr(i,j)$ will be described in detail in the next sub-section.

*C. Calculate the energy (activity correlation) between patches*

The DT energy for the three networks is shown by Fig. 7.

For the scene-related network, since we only need it to detect the movement of the person in our scenario, we simply set all the DT energies to be 1, as in Fig. 7 (a). Note that if we want to detect abnormal group activities such as a group of people following abnormal paths, we can also utilize the method described in Fig. 4 to automatically train the DT energies in the scene-related network instead of simply putting all DT energies to be 1. Furthermore, we can also extend the scene-related network by using directed networks to handle the scenarios related to motion directions (e.g., the road traffic case).

For the normal relative network, three DT energy values are used as shown in Fig. 7 (b). For edges pointing toward the center node, their DT energy values $enr(i,j)$ will be 1 (as the red dashed arrows in Fig. 7 (b)). For edges pointing outward the center node, their DT energy values will be -1 (as the blue dash-dot arrows in Fig. 7 (b)). And the DT energy values will be 0 for edges between nodes having the same distance to the center node (as the black solid arrows in Fig. 7 (b)). Since in the normal relative network, person 1 is fixed at the center node, the normal relative energy $ENR(u_2-u_1,q_2-q_1)$ is mainly calculated by the movement of person 2 with respect to person 1. Based on our DT energy definition, when person 2 is moving close to the center node (i.e., moving toward person 1), $ENR(u_2-u_1,q_2-q_1)$ will be increased. On the contrary, when person 2 is leaving the center node, $ENR(u_2-u_1,q_2-q_1)$ will be decreased. In this way, the relative movement between people can be effectively modeled by the transmission energy.

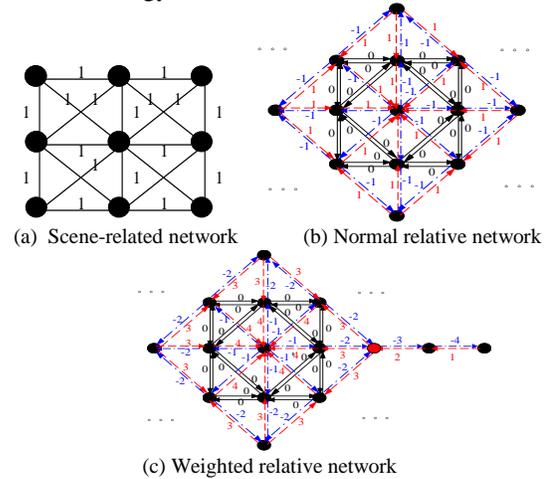

(a) Scene-related network  (b) Normal relative network

(c) Weighted relative network

Fig. 7 The DT energy values for the three networks (note that (a) is an undirected network while (b) and (c) are directed networks).

The structure of the weighted relative network is the same as the normal relative network. However, the DT energy values $ewr(i,j)$ are "weighted" as shown in Fig. 7 (c). For edges either pointing toward or outward the center node, the DT energy values will become larger when they are closer to the center node. Their only difference is that edges pointing toward the center node are positive while edges pointing outward the center node are negative. With this weighted relative network, we can extract the "history" or "temporal" information of the relative movement between people. For example, when person 2 moves from the red node in Fig. 7 (c) toward person 1 and moves back, the corresponding total weighted relative transmission energy $EWR(u_2-u_1,q_2-q_1)$ will be a positive value. On the contrary, $EWR$ will be a negative value when person 2 leaves person 1 from the red node and then comes back.

If we take a more careful look at the three networks in Fig. 7, we can see that since the scene-related network is constructed based on the scene without being affected by the people movements, it can be viewed as an identical field where packages need to consume energy to move and their moving distances are proportional to their consumed energies. Comparatively, since the two relative networks in Fig. 7 (b)-(c) are constructed based on person 1, they can be viewed as the repulsive fields where person 1 in the network center is creating "repulsive" forces. Thus, packages need to consume energy in order to approach person 1 while "gain" energy when leaving person 1. At the same time, no energy will be consumed or gained when packages are revolving around person 1.

Furthermore, note that the three energy networks in Fig. 7 are not fully connected (i.e., each node is only connected to its eight neighboring nodes and are not connected with its non-adjacent nodes). However, these networks can also be extended to become fully connected. For example, we can define the DT energy between two non-adjacent nodes $i$ and $j$ as the minimum possible energy needed to move from $i$ to $j$. In this way, the DT energy between any nodes can be calculated and the fully-connected networks can be constructed.

*D. Activity detection rules*

With the three networks and their corresponding DT energies, we can calculate the total transmission energy set for the input group activity trajectories, as in Eqn. (10). Then when recognizing group activities, we can view the total transmission energy set in Eqn. (10) as a feature vector and train classifiers for automatically achieving the detection rules. In this paper, we use Support Vector Machine (SVM) [8] to learn the detection rules from the training set and use it for group activity recognition. It should be noted that the proposed method can be used with general classifiers. We choose SVM since it is the common choice for activity recognition so that it is easy to be implemented and compared with our methods. Experimental results demonstrate that our NTB algorithm can effectively recognize various group activities.

## V. EXPERIMENTAL RESULTS

In this section, we show experimental results for our proposed NTB algorithm. In the following, we will show the results on four different datasets including abnormality detection and group activity recognition. Finally, we will also discuss the computation complexity and memory storage requirement of our NTB algorithm. Furthermore, note that in our experiments, when we map the trajectory of an object into the patch-based route (such as in Fig. 1 (b)), we check the location of the object in each frame. And as long as the object moves to a new patch, this new patch will be added to the object's patch-based route. In this way, all the patches that the trajectory passes through can be added in the route.

*A. Experiments for abnormal event detection in an abnormality dataset*

First, we perform experiments on our multi-camera dataset. The dataset is created by a two-static-camera system as shown in Fig. 8. From Fig. 8, we can see that the entire room has 5 cubes (the grey blocks) and one entrance door (the dashed block). In normal cases, people enter the room to their cubes, or exit the room from their cubes, or move from one cube to another. Therefore, these cubes and the entrance door can be viewed as the entrance (or exit) patches. Furthermore, two cameras are used to monitor the entire room where one camera monitors the right part (the left blue dashed camera in Fig. 8) and the other one monitors the left part (the right red dash-dot one in Fig. 8).

In total, there are 326 sequences in our dataset which includes 230 normal activity sequences and 96 abnormal activity sequences. Note that each sequence includes two videos from the two cameras. In our experiments, three types of abnormal activities are defined as in Table 1. Fig. 9 shows part of the global trajectories that we extracted for normal activities in the two-camera view where the trajectories from two cameras are first extracted by particle-filter-based method [16] and then combined into the global trajectory by the method of Prosser [6]. Besides, the colored blocks in Fig. 10 are the patches that we divided in our experiment.

Table 1 Three types of abnormal activities in our experiments

| |
|---|
| (I) follow an irregular path (e.g., climb over the discussion table as in Fig. 6 (b)) |
| (II) moving back and forth once or more than once in the room (e.g., in Fig. 6 (c)) |
| (III) approach unusual regions (e.g., go to the left-bottom corner as in Fig. 6 (d)) |

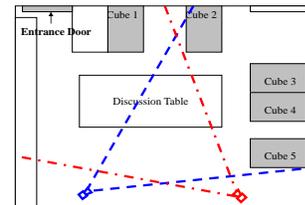

Fig. 8 The configuration of cameras in our two-camera dataset.

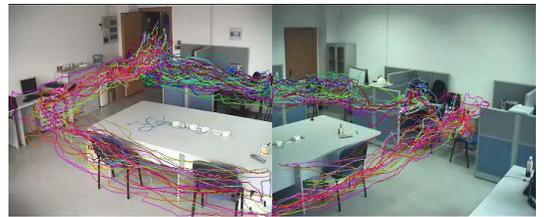

Fig. 9 Global trajectories for part of the regular activities.

Furthermore, several things need to be mentioned about combing multiple camera views: (1) Our proposed algorithm is general and other methods [9, 20-21] can also be used to achieve the activity trajectories; (2) When the patches from different camera views overlap (i.e., patches from different camera views are for the same region), we simply set the DT energy between these patches to be 0 since moving between these patches do not consume any energy; (3) Besides creating the global trajectories, other methods can also be used to combine multi-camera views.

For example, we can first use image stitching [23] to stitch multi-camera images into a large image of the entire scene. Then, we can divide patches on this large image and apply our method.

It should be noted that the scenario of our experiment is quite challenging and complex because: (a) The two-camera view includes both overlapping regions (i.e., regions covered by both cameras) and non-overlapping regions (i.e., regions only covered by a single camera). (b) There are only about 250 sequences available for training (in 75% training and 25% testing case), which is difficult for constructing satisfactory detection models.

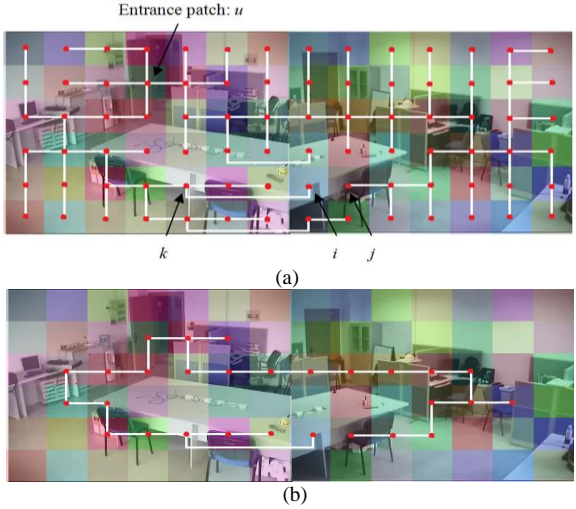

Fig. 10 The minimum-energy route map and normal transmission routes. (a) The minimum-energy route map calculated by our SBIP algorithm. (b) The map for normal transmission routes by deleting connections with large DT values in (a).

Fig. 10 (a) shows the result of a minimum-energy route map calculated by our SBIP algorithm under 75% training and 25% testing where five independent experiments are performed and the results are averaged (note that we train on 75% of all the data where both normal and abnormal samples are included). From this map, we can achieve the minimum-energy routes from the entrance patch to all the patches in the scene. Furthermore, by deleting the connections with large DT values in Fig. 10 (a) (note that large connections with large DT values refer to the routes to the unusual patches), a "normal" route map can be achieved which can be roughly regarded as the map for normal transmission routes, as in Fig. 10 (b). From Fig. 10 (b), we can see that the calculated normal transmission routes among the cubes and the entrance door are pretty close to the regular trajectories in Fig. 9. This implies the effectiveness of our algorithm in detecting abnormal activities. Furthermore, note that: (1) although some patches such as $i$ and $j$ in Fig. 10 (a) are not connected (because the route map only allows one route from the entrance patch to each patch and circle routes are not allowed), it does not mean that the DT energy between $i$ and $j$ is large. Rather, the DT energy between $i$ and $j$ is small. Therefore, during the testing part, people moving downward around the table to patch $j$ can also be detected as normal activity since its total transmission energy is small. (2) Some non-adjacent patches (such as $k$ and $i$ in Fig. 10 (a)) are also connected in the minimum-energy route map. This is because these patches are for the overlap regions (i.e., patches from different camera views but representing the same physical region). And as mentioned, the DT energies between these patches are set to 0 by our algorithm. In this way, patch $i$ can find its best route to the entrance patch $u$ by going through $k$.

Furthermore, Table 2 compares the activity detection results of the following seven methods:

(1) The baseline 1 method which views the patches covered by the normal training trajectories as "normal" patches (i.e., patches went through by the normal training trajectories) and the remaining patches as "abnormal" (or unusual) ones. Thus, in the testing part, trajectories going through those "abnormal" patches will be detected as abnormal (Baseline 1 in Table 2).

(2) The baseline 2 method which uses kernel density estimation (KDE) [24] to construct an occupancy probability based on the normal training trajectories and detects trajectories that enter into low occupancy-probability areas as the abnormal activities. (Baseline 2 in Table 2).

(3) The trajectory-similarity-based method where abnormalities are detected if there is a clear difference between the input trajectory and the pre-trained trajectory cluster [4] (TSB in Table 2).

(4) The spatio-temporal-analysis-based method which first extracts dynamic instants from the global trajectory and then utilizes view-invariant dynamic time warping for measuring trajectory similarities for detection [3] (STAB in Table 2).

(5) The probability-transition-matrix-based method which calculates the activity's conditional probability based on the pre-trained probability transition matrix for activity detection [9] (PTM in Table 2).

(6) The NTB+SVM method. That is, using $[E(u, q), E(u, q)/E_{min}(u, q)]$ as a 2-dimensional feature vector for describing the activities and using SVM to take the place of the rules in Eqn. (7) for abnormity detection (NTB+SVM in Table 2). Note that the training process of the NTB+SVM method is similar to the NTB algorithm as in Fig. 4. However, there are two major differences for the training process of the NTB+SVM method: (a) Since in the abnormal activity recognition scenario, the DT energy values (i.e., $e(i, j)$ which are used to calculated $E(u, q)$ and $E_{min}(u, q)$) also need to be trained in the training process, the SVM classifier needs to be re-trained during each iteration when the DT energy values are updated. That is, the SVM re-train step is used to take the place of threshold updating step (Step 2) in Fig. 4. (b) Since there are no thresholds in the NTB+SVM method, the correlation impact weight $tw_k(i, j)$ can be updated by Eqn. (12) instead of Eqn. (5):

$$tw_k^{n+1}(i,j) = \begin{cases} tw_k^n(i,j) & \text{if k-th trajectory is correctly recognized} \\ tw_k^n(i,j) \cdot (1+P_k^n) & \text{if k is a false alarm} \\ tw_k^n(i,j) \cdot (1-P_k^n) & \text{if k is a miss} \end{cases} \quad (12)$$

where $P_k^n$ is the activity detection probability for the $k$-th trajectory calculated by the SVM in the $n$-th iteration.

(7) Our proposed NTB algorithm (NTB in Table 2).

In Table 2, three rates are compared: false alarm rate (FA) [12], miss detection rate (Miss) [12], and total error rate (TER) [12]. The FA rate is defined by $N_\theta^{fp}/N_\theta^-$ where $N_\theta^{fp}$ is the number of false positive video clips for activity $\theta$ (i.e., the number of normal activities wrongly detected as abnormal activities in this experiment), and $N_\theta^-$ is the total number of negative video clips except activity $\theta$ in the test data (i.e., the total number of normal activities in this experiment) [12]. The miss detection rate is defined by $N_\theta^{fn}/N_\theta^+$ where $N_\theta^{fn}$ is the number of false negative (misdetection) sequences for activity $\theta$ (i.e., the number of abnormal activities wrongly detected as normal activities in this experiment), and $N_\theta^+$ is the total number of positive sequences of activity $\theta$ in the test data (i.e., the total number of abnormal activities in this experiment) [12]. The TER rate is calculated by $N_{t\_r}/N_{t\_f}$ where $N_{t\_r}$ is the total number of wrongly detected activities for both normal and abnormal activities and $N_{t\_f}$ is the total number of activity sequences in the test set. TER reflects the overall performance of the algorithm in detecting both the normal and the abnormal activities [12]. In order for more detailed comparison, we also include the Miss rate for each individual abnormal activity listed in Table 1 (i.e., the miss rates for I, II, III in Table 2). Note that since the two baseline methods (Baseline 1 and Baseline 2) are only designed to detect normal and abnormal activities, they cannot differentiate different abnormality types and thus the miss rates for type I, II, III abnormalities for these two methods are not listed in Table 2. Furthermore, also note that the abnormal activities can be simply differentiated in our NTB algorithm by: (a) detecting as type I abnormality if the activity's total transmission energy (TE) is larger than both thresholds $T_1$ and $T_2$, (b) detecting as type II abnormality if TE is smaller than $T_1$ and larger than $T_2$, and (c) detecting as type III abnormality if TE is larger than $T_1$ but smaller than $T_2$. Some examples of the process for detecting these three abnormality types are shown in Fig. 6.

Table 2 Miss, FA and TER rates of abnormal activity detection under 75% training and 25% testing

|  |  | Baseline 1 | Baseline 2 | TSB | STAB | PTM | NTB+SVM | NTB |
|---|---|---|---|---|---|---|---|---|
| FA (%) |  | 20.3 | 17.4 | 19.0 | 17.9 | 11.9 | **3.7** | 3.6 |
| Miss (%) | (I) | - | - | 33.3 | 23.8 | 28.6 | **8.8** | 9.6 |
|  | (II) | - | - | 85.7 | 71.4 | 66.7 | **13.0** | 14.3 |
|  | (III) | - | - | 42.9 | 38.1 | 47.6 | **27.3** | 28.6 |
|  | Total | 40.0 | 36.9 | 28.6 | 20.6 | 27.0 | **15.1** | 15.4 |
| TER (%) |  | 30.8 | 27.3 | 23.1 | 19.0 | 18.4 | **9.1** | 9.3 |

From Table 2, we can see that the performance of our NTB algorithm is obviously better than the other methods (Baseline 1, Baseline 2, TSB, STAB, and PTM). Besides, several observations can be drawn from Table 2.

(1) The performance of our NTB algorithm is obviously better than the two baseline methods. This is because: (a) Since the training data in this experiment are not sufficient, "normal" regions are not fully covered by the normal training trajectories (i.e., some regions are normal but no normal training trajectory passes through them). Thus, if we simply use the limited normal training data to model all the normal routes (such as the two baseline methods), many normal regions will be mis-regarded as the "abnormal" ones and the detection performance will be greatly affected. Comparatively, our NTB algorithm utilizes an iterative method to construct the DT energies between patches by suitably integrating both the normal and abnormal training samples as well as the error rates on these training samples (i.e., Eqns (5) and (9) in the paper). In this way, the insufficient training data can be more efficiently utilized to construct a more reliable model. (b) More importantly, the two baseline methods also cannot differentiate the abnormalities whose entire trajectories are inside the normal regions (e.g., moving back and forth in the regular route or moving around the table in the regular route). Comparatively, our method can effectively detect these abnormalities by checking the second criteria in Eqn. (7).

(2) Our NTB algorithm also has better performance than the trajectory-similarity-based methods such as TSB and STAB. This is because: (a) the trajectory-similarity-based methods will easily confuse large-deviation normal trajectories with small-deviation abnormal trajectories. For example, if a normal trajectory keeps zigzagging around the normal route, its distance to the normal-route cluster may be large. At the same time, if an abnormal trajectory closely follows the normal route most of the time but only deviates to unusual region at the end, its distance to the normal-route cluster may be even smaller than the normal trajectory. In this case, the trajectory-similarity-based methods will fail to work. Comparatively, our NTB algorithm utilizes both the normal and abnormal training samples as well as the error rates on these training samples to construct suitable DT energies between patches. In this way, trajectories moving into unusual regions will be effectively detected due to the large DT energies entering these unusual patches. (b) The trajectory-similarity-based methods also have low efficiency in detecting abnormalities such as back and forth whose trajectory overlaps. Comparatively, our NTB algorithm can work effectively by checking the second criteria in Eqn. (7).

(3) Although the PTM method introduces the transmission matrix and multi-camera consensus for handling activity detection, its performance is still less satisfactory. This is because: (a) The transmission matrix may have limitations in differentiating different trajectory behaviors. (b) The camera overlapping area in this experiment is small, which limits the capabilities of its multi-camera consensus step. On the contrary, our NTB algorithm is more flexible under this scenario.

(4) Our NTB algorithm is not only effective in detecting abnormal activities, but also efficient in differentiating all the abnormality activity types (i.e., I, II, III). Compared to our method, the abnormal activity differentiation ability for the other methods are much poorer. Furthermore, the compared methods (i.e., TSB, STAB, PTM) are extremely poor in differentiating the "back and forth" activity (i.e., (II)). This is because many "back" trajectory patches are overlapped with the "forth" trajectory patches, making the entire trajectory very difficult to be efficiently represented. However, this type of activities can still be effectively detected by NTB as shown in Fig. 6 (b).

(5) Comparing NTB with NTB+SVM methods, we have seen that both methods can achieve similar performances. This demonstrates that: (a) our proposed energy-based features ($E(u, q)$ and $E_{min}(u, q)$) are effective in differentiating abnormal activities; (b) Our energy-based features from the network models are general and other general classifiers can also be utilized to perform abnormal activity recognition besides the criteria in Eqn. (7).

Furthermore, Fig. 11 shows the total transmission energies calculated by our NTB algorithm for one set of testing sequences where the first 34 values are for normal activity sequences and the later 21 values are for abnormal activity sequences. We can see from Fig. 11 that most abnormal activities have large total energy values by our model and thus can be easily detected. Some abnormal activities have relatively small absolute energy values. For example, sequences 39-42 correspond to activities going back and forth along the normal path. Since most of their trajectories are on the 'normal' route where the DT energies are small, the accumulated total energies for these sequences become small. However, since their energy differences with the minimum possible energy $E_{min}$ are larger than our pre-trained threshold, they still can be successfully detected in our algorithm by checking the second criterion in Eqn. (7).

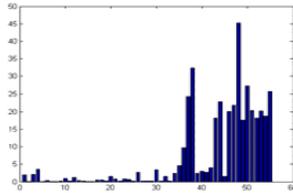

Fig. 11 The total transmission energies calculated by our NTB algorithm for activities in the test sequences.

Table 3 Miss, FA and TER rates of the detection algorithms with different patch sizes (Note that the results in Table 2 and Fig. 10 are achieved by using the patch size of 48×48)

| Patch Size | 9×9 | 24×24 | 32×32 | 36×36 | 41×41 | 48×48 | 72×72 |
|---|---|---|---|---|---|---|---|
| TER | 18.3 | 10.8 | 10.2 | 8.9 | 9.5 | 9.3 | 17.2 |

Finally, we also perform another experiment by using different patch sizes for abnormality detection. The results are shown in Table 3 and we can have the following observations.

(1) We can achieve stable results when the patch sizes change within a wide range (e.g., from 24×24 to 48×48 in Table 3). This implies that the iterative training method in our NTB algorithm can adaptively achieve suitable DT energy values for different patch sizes when the patch size is within a reasonable range.

(2) The patch size cannot be extremely large. When the patch size is extremely large (e.g., 72×72 in Table 3), there will be few patches in the scene. This will make the algorithm difficult to differentiate the various activity patterns. For example, in the extreme case, if the patch size is the entire image and there is only one patch, it is impossible to perform recognition.

(3) Also, the patch size also cannot be extremely small. When the patch size is extremely small (e.g., 9×9 in Table 3), the number of nodes and edges in the network will become obviously large. In this case, large number of training samples is required in order for constructing reliable DT energies. Otherwise, the performance will be poor. For example, if the patch size is 1×1 (i.e., each pixel is a patch) and we only have 10 training trajectories, there will be large number of "normal" patches where no training trajectory arrives. In this case, the DT energy for these patches will be large and trajectories passing these patches will be easily detected as abnormal.

(4) From the above discussions, in the experiments in our paper, we select patch sizes such that the entire image of one camera scene can have 7-14 patches in width and 6-12 patches in height. Of course, when more training samples are available, smaller patch sizes can also be selected.

*B. Experiments for group activity recognition on the BEHAVE dataset*

We further perform another set of experiments for the group activity recognition. The experiments are performed on the public BEHAVE dataset [18] where 800 activity clips are selected for recognition. Eight group activities are recognized as shown in Table 4. Some frames are shown in Fig. 12.

Table 5 compares the results of the four methods:
(1) The group-representative-based algorithm [12] (GRAD in Table 5).
(2) The pair-activity classification algorithm based on bi-trajectories analysis which uses causality and feedback ratios as features [13] (PAC in Table 5).
(3) The localized-causality-based algorithm using individual, pair, and group causalities for group activity detection [14] (LCC in Table 5).
(4) Our proposed NTB algorithm with transmission energy sets from three networks (NTB in Table 5).

Table 4 The group activities recognized on the BEHAVE dataset

| (I) meet: two people walk toward each other. |
|---|
| (II) follow: two people are walking. One people follow another. |
| (III) approach: one people stand and another walk toward the first people. |
| (IV) separate: two people escape from each other. |
| (V) leave: one people stand and another leave the first people. |
| (VI) together: two people are walking together. |
| (VII) exchange: two people first gathered and then leave each other. |
| (VIII) return: two people first separate and then meet. |

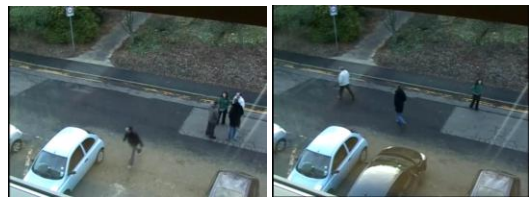

(a) Leave  (b) Follow
Fig. 12 Example frames of the BEHAVE dataset.

Table 5 Miss, False Alarm, and TER rates of the group recognition algorithms under 75% training and 25% testing

|  |  | GRAD | PAC | LCC | **NTB** |
|---|---|---|---|---|---|
| Meet | Miss(%) | 11.4 | 12.1 | 18.5 | **0.0** |
|  | FA(%) | 1.6 | 1.3 | 2.4 | **0.0** |
| Follow | Miss(%) | 12.2 | 8.3 | 11.7 | **2.5** |
|  | FA(%) | 0.9 | 0.8 | 1.4 | **0.2** |
| Approach | Miss(%) | 10.3 | 9.0 | 12.8 | **0.0** |
|  | FA(%) | 1.8 | 2.6 | 2.6 | **0.5** |
| Separate | Miss(%) | 6.3 | 5.0 | 5.8 | **2.5** |
|  | FA(%) | 1.2 | 0.7 | 1.9 | **0.6** |
| Leave | Miss(%) | 5.2 | 4.9 | 9.7 | **5.6** |
|  | FA(%) | 1.8 | 1.1 | 1.4 | **1.0** |
| Together | Miss(%) | 2.7 | 2.9 | 6.6 | **8.8** |
|  | FA(%) | 0.7 | 0.2 | 0.9 | **0.9** |
| Exchange | Miss(%) | 24.7 | 28.9 | 26.3 | **2.6** |
|  | FA(%) | 2.4 | 3.2 | 2.7 | **0.5** |
| Return | Miss(%) | 28.6 | 36.8 | 31.6 | **5.3** |
|  | FA(%) | 2.6 | 2.5 | 2.3 | **0.2** |
| TER(%) |  | 10.7 | 11.1 | 13.8 | **3.4** |

In Table 5, three rates are compared: the miss detection rate (Miss), the false alarm rate (FA) [12], and the total error rate (TER). From Table 5, we can see that our proposed NTB algorithm can achieve obviously better performance than the other three state-of-art algorithms. This demonstrates that our NTB algorithm with the transmission energy features can precisely catch the inter-person spatial interaction and the activity temporal history characteristics of the group activities. Specifically, our NTB is obviously effective in recognizing complex activities (i.e., exchange and return). In Fig. 13, (a) shows two example trajectories of the complex activities, (b) shows the values of the major features in the PAC algorithm [13], and (c) shows the transmission energy (EWR) from the weighted relative network in our NTB algorithm. From Fig. 13 (b), we can see that the features in the PAC algorithm [13] cannot show much difference between the two complex activities. Compared to (b), our EWR energy in (c) is obviously more distinguishable by effectively catching the activity history information.

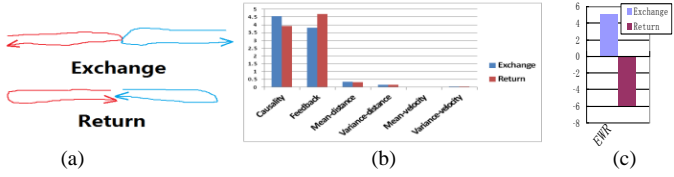

Fig. 13 (a) Example trajectories for complex group activities; (b) The major feature values for PAC algorithm; (c) The EWR energy values by our NTB.

### C. Experiments for group activity recognition on the CASIA dataset

In order to further demonstrate the effectiveness of our NTB algorithm, we also perform another experiment on the CASIA dataset [26]. The CASIA dataset contains seven group activities as shown in Table 6 [26]. Some example frames are in Fig. 2.

Table 6 The group activities recognized on the CASIA dataset

| |
|---|
| $A_1$ (**rob**): person $P_1$ follows person $P_2$, catches him, robs him, and then runs away. |
| $A_2$ (**fight**): people $P_1$ and $P_2$ approach each other and fight with each other. |
| $A_3$ (**follow**): person $P_1$ follows person $P_2$ until the end. |
| $A_4$ (**follow and gather**): person $P_1$ follows person $P_2$ and then walks together |
| $A_5$ (**meet and part**): $P_1$ and $P_2$ approach each other, meet, and then depart. |
| $A_6$ (**meet and gather**): $P_1$ and $P_2$ meet each other and then walk together |
| $A_7$ (**overtake**): person $P_1$ overtakes person $P_2$. |

Since the activities such as "rob" and "fight" are related to person's local motion intensities, many of the algorithms on this dataset [27] utilize both the trajectory and the motion intensity features for detection. Therefore, in order to have a fair comparison with these methods, we further extend our algorithm by introducing an additional motion-intensity network to include the motion intensity feature. The structure of the motion-intensity network is the same as the scene-related network in Fig. 7 (a). However, different from the scene-related network whose DT energies are a constant value, the DT energies in the motion-intensity network are decided by the motion intensities [27] when an object moves across patches:

$$emi(i, j, t) = s(i, j, t) \quad (13)$$

where $emi(i, j, t)$ is the DT energy between patches $i$ and $j$ at time $t$ in the motion-intensity network. $s(i, j, t)$ is the motion-intensity in patches $i$ and $j$ at time $t$ and it can be calculated by:

$$s(i, j, t) = \left| v_{opticalflow}(i, j, t) - v(i, j, t) \right| \quad (14)$$

where $v_{opticalflow}(i, j, t)$ is the magnitude of the average optical flow speed inside patches $i$ and $j$ at time $t$. And $v(i, j, t)$ is the magnitude of the object's global speed moving across patches $i$ and $j$ at time $t$. Similar to [27], $v_{opticalflow}(i, j, t)$ and $v(i, j, t)$ can be calculated from the Lucas–Kanade algorithm [28] and the object's trajectory, respectively [27].

Basically, since $v_{opticalflow}$ includes both the object's local and global motions while $v$ only includes the global motion, by removing $v$ from $v_{opticalflow}$, the object's local motion intensities can be achieved [27]. Furthermore, note that the DT energy $emi(i, j, t)$ is related to time $t$. This means that people moving across patches with different local motion patterns will have different DT energies in the motion-intensity network. With this motion-intensity network, the following feature vector is utilized in our NTB algorithm for group activity recognition.

$$[E_1(u_1,q_1), E_2(u_2,q_2), ENR(u_2-u_1,q_2-q_1), EWR(u_2-u_1,q_2-q_1),$$
$$EMI_1(u_1,q_1), EMI_2(u_2,q_2)] \quad (15)$$

where the definitions of $E_1(u_1,q_1)$, $E_2(u_2,q_2)$, $ENR(u_2-u_1,q_2-q_1)$, and $EWR(u_2-u_1,q_2-q_1)$ are the same as in Eqn. (10). $EMI_1(u_1,q_1)$ and $EMI_2(u_2,q_2)$ are the total transmission consumption for person 1 and person 2 in the motion-intensity network.

Fig. 14 compares the experimental confusion metric results of different methods on the CASIA dataset. In Fig. 14, (a)-(e) shows the results for the Hidden Markov Model (HMM) method [12, 29], the Coupled HMM method (CHMM) [30], the coupled observation decomposed HMM method with continuous features (CODHMM_C) [27], the coupled observation decomposed HMM method with some discretized features [27], and our NTB algorithm, respectively. From Fig. 14, we can see that our proposed NTB algorithm can also achieve better performance than the state-of-art algorithms [27] on the CASIA dataset. More specifically, our algorithm has obvious improvements in detecting activities such as rob ($A_1$), follow ($A_3$), and overtake ($A_7$). This further demonstrates that: (a) our network-based models are very effective in differentiating similar activities (such as follow and overtake); (b) Besides trajectories, our algorithm can also be extended to include other motion features (such as local motion intensities) and effectively handle the complicated activities such as rob. This point will be further discussed in the next sub-section.

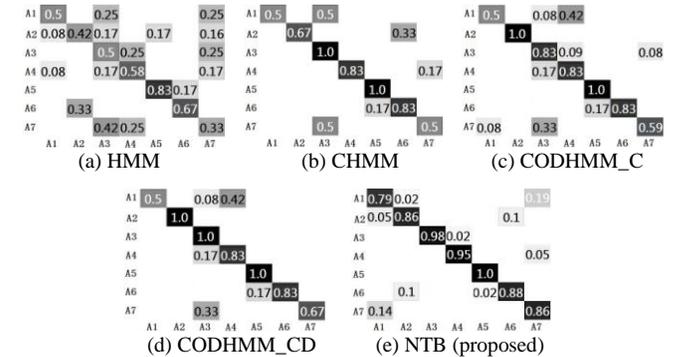

Fig. 14 Confusion matrices for different methods on the CASIA dataset.

### D. Experiments for group abnormality detection on the UMN dataset

In this section, we perform another experiment on the UMN dataset [31] which contains videos of 11 different scenarios of an abnormal "escape" event in 3 different scenes including both indoor and outdoor. Each video starts with normal behaviors and ends with the abnormal behavior (i.e., escape). Some example images of the UMN dataset are shown in Fig. 15.

In order to recognize the abnormal "escape" events, we simply use a single normal relative network (as in Fig. 7 (b)) and put it in the center of the image scene (note that in this experiment, the normal relative network is fixed at the image center rather than moving with some object). Furthermore, instead of extracting the object trajectories, we directly extract the optical flows [27] from the videos and use them as the packages to transmit in the normal relative networks. When detecting activities, we use a sliding window to segment the video into small video clips [12] and the total transmission energy of all optical-flow packages in the video clip is used to detect events in this video clip (i.e., we simply compare the total transmission energy with a threshold to detect the abnormal "escape" events).

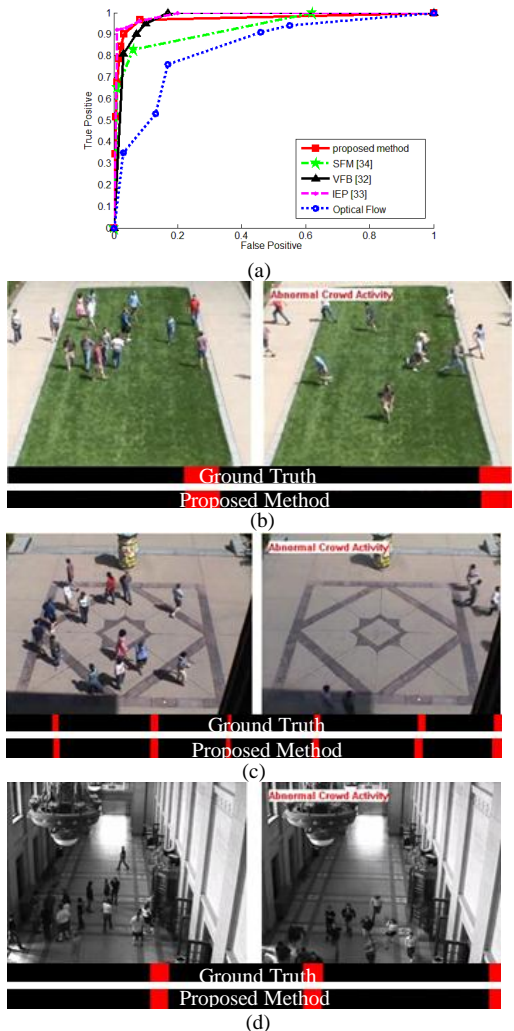

Fig. 15 (a) ROC results of different methods on the UMN dataset; (b)-(d) The qualitative results of using our NTB algorithm for abnormal detection in the UMN dataset. The bars represent the labels of each frame, black represents normal and red represents abnormal. (best viewed in color)

Fig. 15 (b)-(d) compares the normal/abnormal classification results of our algorithm with the ground truth. Furthermore, Fig. 15 (a) compares the ROC curves between our algorithm (Proposed) and the other four algorithms: the optical flow only method (Optical Flow) [12, 32], the Social Force Model (SFM) [34], the Interaction Energy Potential method (IEP) [33], and the Velocity-Field Based method (VFB) [32]. The results in Fig. 15 (a) show that by using the simple optical flow feature and a single relative network, our algorithm can achieve similar or better results than the state-of-the-art algorithms [32-34]. This further demonstrates the effectiveness of our NTB algorithm. Furthermore, several things need to be mentioned about Fig. 15. (1) Basically, in our normal relative network (as in Fig. 7 (b)), packages moving to different directions will cause different positive and negative energies. Therefore, in normal events, since people walk randomly in various directions, the optical-flow packages will create similar amounts of positive and negative energies such that the total transmission energy will be around zero. However, during the abnormal "escape" events, since people are coherently moving "outside", most optical-flow packages will create negative energies. Thus, the total transmission energy will be a large negative number and the abnormalities can be effectively detected.

(2) In this experiment, the low-level optical flow features are used. This implies that in cases when reliable trajectory cannot be achieved (e.g., in extremely crowded scenes), we can also extend our algorithm by using the low-level motion features to take the place of the trajectories for recognition.

(3) Note that in this experiment, we only use the basic optical flow feature and a single relative network for detection. The performance of our NTB may be further improved by: (a) using more reliable motion features such as the improved optical flow [23], (b) introducing additional networks to handle more complicated scenarios.

### E. Computation complexity and memory storage requirements

Finally, we evaluate the computation complexity and memory storage requirements of our algorithm.

**(1) Computation complexity.** Our algorithm is run on a PC with 2.6 GHz 2-Core CPU and 4 G RAM while the training and testing processes are implemented by Matlab.

Table 7 Computation costs of our NTB algorithm on different datasets

|  | Abnormality dataset | | BEHAVE | CASIA | UMN |
|---|---|---|---|---|---|
|  | NTB+SVM | NTB |  |  |  |
| Learning | 10 min | 6 min | 50 sec | 10 sec | <10 ms |
| Evaluation | 10 ms | 10 ms | 10 ms | <20 ms | <10 ms |

Table 7 shows the computation costs of our NTB algorithm in the experiments of Sections V-A to V-D. From Table 7, we can see that for the abnormality detection dataset, given about 250 training trajectories, the training process took about 6 minutes to converge. And the entire testing process only took about 10 ms to process over 80 input trajectories since we only need to calculate the transmission energies by applying Eqn. (1). Besides, when combined with human detector [20] and particle-based tracking [6, 16] (implemented by C++), our algorithm can still achieve about 20 frames/sec in the testing process. Therefore, our algorithm has low computation complexity and is suitable for real-time applications.

For group activity recognition in the BEHAVE dataset, given about 600 group trajectories in the training set, the training process took about 50 seconds since the DT energies in the group activity experiments do not need to be trained in an iterative way. And entire testing processing also only took about 10 ms to process over 200 input trajectories. Similarly, our algorithm's complexity cost on the CASIA and UMN datasets are also low.

Moreover, Table 8 compares the computation costs of our algorithm with the other methods on the CASIA dataset [27] (note that the complexity costs of the other methods are achieved from their publications [27] in order for a fair comparison). From Table 8, we can also see that the computation complexity of our algorithm is lower than the compared methods.

Table 8 Computation costs of different methods on the CASIA dataset

|  | HMM | CHMM | CODHMM_C | CODHMM_CD | Proposed |
|---|---|---|---|---|---|
| Learning | <1 min | <1 min | 1 min | 1 min | 10 sec |
| Evaluation | 20 ms | 30 ms | 30 ms | 30 ms | <20 ms |

**(2) Memory storage requirements.** As for the storage issue, for abnormal event detection case, we need an $N \times N$ matrix to store the fully connected network (i.e., the $N \times N$ DT energies between nodes where $N$ is the total number of nodes) and an $N \times 1$ matrix to store the minimum possible energy $E_{min}$ from the entrance patch $u$ to all the other patches. In the example of Figs 9-10, we have totally 84 nodes and thus only one $84 \times 84$ and one $84 \times 1$ float type matrixes are needed, which is small load for memory. For group activity recognition case, since the networks in Fig. 7 are pre-calculated. We even do not need matrixes to store the DT values and all the DT energy values can be derived according to the relative location between people. Thus, the storage requirements of the proposed method are low.

## VI. CONCLUSION AND FUTURE WORK

In this paper, a new network-based algorithm is proposed for human activity recognition in videos. The proposed algorithm models the entire scene as a network. Based on this network, we further model people in the scene as packages. Thus, various human activities can be modeled as the process of package transmission in the network. By analyzing the transmission process, various activities such as abnormal activities and group activities can be effectively recognized. Experimental results demonstrate the effectiveness of our algorithm.

In the future, our algorithm may be further extended in the following ways: (a) In this paper, we assume that the camera is fixed and directly divide patches in the image of the scene. However, we can extend our algorithm by setting up a global coordinate of the entire scene and divide patches of the scene in this global coordinate. In this way, even when the camera is moving or zoom-in/zoom-out, we can also handle these cases by first mapping the person's location into this global coordinate [25] and then performing our algorithm inside the global coordinate. (b) Although our algorithm can perform online detection, this online detection capability will finish when the abnormality is detected (i.e., after the abnormality is detected, all the later parts will be detected as abnormal). However, note that our algorithm can be extended to further handle the online detection task even after abnormality happens. For example, we can use the sliding windows to segment the video into clips and then perform detection in each video clip independently. And these will be one of our future works.

## REFERENCES

[1] J. K. Aggarwal and M. S. Ryoo, "Human Activity Analysis: A Review," *ACM Computing Surveys (CSUR)*, vol. 43, no. 16, pp. 1-47, 2011.
[2] J. Nascimento, M. Figueiredo, and J. Marques, "Segmentation and classification of human activities," *Int'l Workshop Human Activity Recognition and Modeling*, pp. 79-86, 2005.
[3] C. Rao, M. Shah, T. Syeda-Mahmood, "Action recognition based on view invariant spatio-temporal analysis," *Proceedings of ACM Multimedia,* pp. 518-527, 2003.
[4] E. E. Zelniker, S. Gong and T. Xiang, "Global abnormal behavior detection using a network of CCTV cameras," *Int'l Workshop. Visual Surveillance*, pp. 1-8, 2008.
[5] Jaechul Kim and Kristen Grauman, "Observe Locally, Infer Globally: a Space-Time MRF for Detecting Abnormal Activities with Incremental Updates," *IEEE Conf. Computer Vision and Pattern Recognition (CVPR)*, pp. 2921-2928, 2009.
[6] B. Prosser, S. Gong and T. Xiang, "Multi-camera matching under illumination change over time," *Workshop on Multi-camera and Multi-modal Sensor Fusion Algorithms and Applications*, pp. 1-12, 2008.
[7] C. C. Loy, T. Xiang and S. Gong. "Modelling activity global temporal dependencies using time delayed probabilistic graphical model," *Int'l Conf. Computer Vision (ICCV)*, pp 120-127, 2009.
[8] C.-C. Chang and C.-J. Lin, "LIBSVM : a library for support vector machines," *ACM Trans. Intelligent Systems and Technology*, vol. 2, no. 3, pp. 1-27, 2011.
[9] B. Song, A. Kamal, and C. Soto, "Tracking and activity recognition through consensus in distributed camera networks," *IEEE Trans. Image Processing,* vol. 19, pp. 2564-2579, 2010.
[10] D. Zhang, D. Gatica-Perez, S. Bengio, and I. McCowan, "Modeling individual and group actions in meetings with layered HMMs," *IEEE Trans. Multimedia*, vol. 8, no. 3, pp. 509–520, 2006.
[11] Z. Cheng, L. Qin, Q. Huang, S. Jiang, and Q. Tian, "Group activity recognition by Gaussian process estimation," *Int'l Conf. Pattern Recognition (ICPR)*, pp. 3228-3231, 2010.
[12] [12] W. Lin, M.-T. Sun, R. Poovendran and Z. Zhang, "Group event detection with a varying number of group members for video surveillance," *IEEE Trans. Circuits and Systems for Video Technology*, pp. 1057-1067, 2010.
[13] Y. Zhou, S. Yan and T. Huang, "Pair-activity classification by bi-trajectory analysis," *IEEE Conf. Computer Vision Pattern Recognition (CVPR)*, pp. 1-8, 2008.
[14] B. Ni, S. Yan and A. Kassim, "Recognizing human group activities with localized causalities," *IEEE Conf. Computer Vision and Pattern Recognition (CVPR)*, pp. 1470-1477, 2009.
[15] J. Li, S. Gong, and T. Xiang, "Scene segmentation for behavior correlation," *ECCV*, pp. 383–395, 2008.
[16] R. Hess and A. Fern, "Discriminatively Trained Particle Filters for Complex Multi-Object Tracking," *IEEE Conf. Computer Vision and Pattern Recognition (CVPR)*, pp. 240-247, 2009.
[17] J. Li, S. Gong, and T. Xiang, "Discovering multi-camera behaviour correlations for on-the-fly global prediction and anomaly detection," *Int'l Workshop. Visual Surveillance*, pp. 1330-1337, 2009.
[18] BEHAVE set: http://groups.inf.ed.ac.uk/vision/behavedata/interactions/
[19] J. E. Wieselthier, G. D. Nguyen, and A. Ephremides, "On the construction of energy-efficient broadcast and multicast trees in wireless networks," *INFOCOM*, pp. 586-594, 2000.
[20] J. Wu, C. Geyer and J. M. Rehg, "Real-time human detection using contour cues," *IEEE Conf. Robotics and Automation (ICRA)*, pp. 860-867, 2011.
[21] Z. Kalal, J. Matas, and K. Mikolajczyk, "Online learning of robust object detectors during unstable tracking," *IEEE On-line Learning for Computer Vision Workshop (OLCV)*, pp. 1417-1424, 2009.
[22] G. Wu, Y. Xu, X. Yang, Q. Yan, K. Gu, "Robust object tracking with bidirectional corner matching and trajectory smoothness algorithm," *IEEE International Workshop on Multimedia Signal Processing (MMSP)*, pp. 294-298, 2012.
[23] M. Brown, D. Lowe, "Automatic panoramic image stitching using invariant features," *International Journal of Computer Vision*, vol. 74, no.1, pp. 59-73, 2007.
[24] KDE Toolbox: http://www.ics.uci.edu/~ihler/code/
[25] L. G. Mirisola, J. Dias, and A. T Almeida, "Trajectory recovery and 3d mapping from rotation-compensated imagery for an airship," *IEEE Int'l Conf. Intelligent Robots and Systems,* pp. 1908-1913, 2007.
[26] CASIA Action Database [Online]. Available: http://www.cbsr.ia.ac.cn/english/Action%20Databases%20EN.asp
[27] P. Guo, Z. Miao, X. Zhang, Y. Shen and S. Wang, "Coupled observation decomposed hidden markov model for multiperson activity recognition," *IEEE Trans. Circuits and Systems for Video Technology*, vol. 22, no. 9, pp. 1306-1320, 2012.
[28] B. D. Lucas and T. Kanade, "An iterative image registration technique with an application to stereo vision," *Proceedings of Imaging Understanding Workshop*, pp. 674-679, 1981.
[29] L. McCowan, D. Gatica-Perez, S. Bengio, G. Lathoud, M. Barnard, and D. Zhang, "Automatic analysis of multimodal group actions in meetings," *IEEE Trans. Pattern Anal. Mach. Intell.*, vol. 27, no. 3, pp. 305-317,

2005.
[30] M. Brand, N. Oliver, and A. Pentland, "Coupled hidden Markov models for complex action recognition," in *Proc. IEEE Conf. Comput. Vis. Pattern Recognit.*, pp. 994–999, 1997.
[31] Unusual Crowd Activity Dataset: http://mha.cs.umn.edu/Movies/Crowd-Activity-All.avi
[32] J. Zhao, Y. Xu, X. Yang, and Q. Yan, "Crowd Instability Analysis Using Velocity-Field Based Social Force Model," *Visual Communications and Image Processing (VCIP)*, pp. 1-4, 2011.
[33] X. Cui, Q. Liu, M. Gao, Metaxas and D. N, "Abnormal Detection Using Interaction Energy Potentials," *IEEE Conf. Computer Vision and Pattern Recognition (CVPR)*, pp. 3161-3167, 2011.
[34] R. Mehran, A. Oyama, and M. Shah, "Abnormal Crowd Behavior Detection using Social Force Model," *IEEE Conf. Computer Vision and Pattern Recogntion (CVPR)*, pp. 935-942, 2009.